\documentclass[sigconf]{acmart}
\usepackage{subcaption}
\usepackage{tabularx}
\AtBeginDocument{%
  }

\setcopyright{acmlicensed}
\copyrightyear{2024}
\acmYear{2024}
\acmDOI{XXXXXXX.XXXXXXX}

\acmConference[Conference acronym 'XX]{Make sure to enter the correct
  conference title from your rights confirmation emai}{June 03--05,
  2018}{Woodstock, NY}
\acmISBN{978-1-4503-XXXX-X/18/06}

\copyrightyear{2025}
\acmYear{2025}
\setcopyright{acmlicensed}\acmConference[PEARC '25]{Practice and Experience in Advanced Research Computing}{July 20--24, 2025}{Columbus, OH, USA}
\acmBooktitle{Practice and Experience in Advanced Research Computing (PEARC '25), July 20--24, 2025, Columbus, OH, USA}
\acmDOI{10.1145/3708035.3736003}
\acmISBN{979-8-4007-1398-9/2025/07}

\begin{document}

\title{GreenCrossingAI: A Camera Trap/Computer Vision Pipeline for Environmental Science Research Groups}

\author{Bernie Boscoe}
\email{boscoeb@sou.edu}
\orcid{0000-0001-6790-7297}
\affiliation{%
  \institution{Southern Oregon University}
  \city{Ashland}
  \state{Oregon}
  \country{USA}
}
\author{Shawn E. Johnson}
\email{johnsos32@sou.edu}
\affiliation{%
  \institution{Southern Oregon University}
  \city{Ashland}
  \state{Oregon}
  \country{USA}
}
\author{Andrea J. Osbon}
\email{osbona@sou.edu}
\affiliation{%
  \institution{Southern Oregon University}
  \city{Ashland}
  \state{Oregon}
  \country{USA}
}

\author{Chandler Campbell}
\email{campbellr@sou.edu}
\affiliation{%
    \institution{Southern Oregon University}
    \city{Ashland}
    \state{Oregon}
    \country{USA}
}

\author{Karen H. Mager}
\email{magerk@sou.edu}
\affiliation{%
  \institution{Southern Oregon University}
  \city{Ashland}
  \state{Oregon}
  \country{USA}
}

\renewcommand{\shortauthors}{Boscoe et al.}

\begin{abstract}
  Camera traps have long been used by wildlife researchers to monitor and study animal behavior, population dynamics, habitat use, and species diversity in a non-invasive and efficient manner. While data collection from the field has increased with new tools and capabilities, methods to develop, process, and manage the data, especially the adoption of ML/AI tools, remain challenging. These challenges include the sheer volume of data generated, the need for accurate labeling and annotation, variability in environmental conditions affecting data quality, and the integration of ML/AI tools into existing workflows that often require domain-specific customization and computational resources. This paper provides a guide to a low-resource pipeline to process camera trap data on-premise, incorporating ML/AI capabilities tailored for small research groups with limited resources and computational expertise. By focusing on practical solutions, the pipeline offers accessible approaches for data transmission, inference, and evaluation, enabling researchers to discover meaningful insights from their ever-increasing camera trap datasets.

\end{abstract}

\ccsdesc[300]{Computing methodologies~Object detection}
\ccsdesc[300]{Computing methodologies~Object recognition}
\ccsdesc[300]{Computing methodologies~Neural networks}
\ccsdesc[300]{Computing methodologies~Supervised learning by classification}
\ccsdesc[300]{Applied computing~Life and medical sciences}
\ccsdesc[300]{Information systems~Data mining}
\ccsdesc[500]{Software and its engineering~Software development methods}

\keywords{Environmental Science, MegaDetector, Camera Traps}

\maketitle

\section{Introduction}
Wildlife researchers increasingly rely on camera traps to collect large-scale ecological data, enabling detailed studies of species behavior, population trends, and habitat utilization. While the proliferation of camera traps and advancements in machine learning (ML) tools like MegaDetector to detect animals in images have made it possible to partially automate image processing tasks, integrating these tools into practical workflows remains a significant challenge for many research groups. This is especially true for smaller teams with limited computational resources and expertise.
Existing solutions for camera trap data processing, including cloud-based platforms and custom pipelines, often require significant infrastructure, technical know-how, and funding. These barriers leave smaller research groups dependent on manual processing methods, which are time-consuming and error-prone, or unable to fully utilize the data they collect. 
As a result, there is a pressing need for accessible, low-resource solutions that bridge the gap between cutting-edge ML tools and practical usability in ecological research.

This paper presents the GreenCrossingAI project, which developed an on-premise data processing pipeline tailored for camera trap data for a small Environmental Science research group. The pipeline is designed to address common challenges, such as transferring large volumes of data, automating image analysis with machine learning tools like MegaDetector, and efficiently storing and managing results. By leveraging small-scale hardware and open source software, the system provides a scalable solution for research groups without advanced computational infrastructure.
The following sections describe the GreenCrossingAI project in detail, starting with its origins in a wildlife crossing project involving Interstate 5 in southern Oregon, camera trap analysis system design, pipeline workflow, and implementation results. While this pipeline includes a computer vision algorithm to detect animals in camera trap images, here we do not evaluate the accuracy of the model on the researcher’s data, instead we focus on the implementation strategies, and meeting the computing needs of the researchers.  This work demonstrates that even small research teams can harness the power of ML for ecological studies, advancing both scientific research and computational accessibility.
\begin{figure*}
  \includegraphics[trim={0 45 0 0},clip,width=\textwidth]{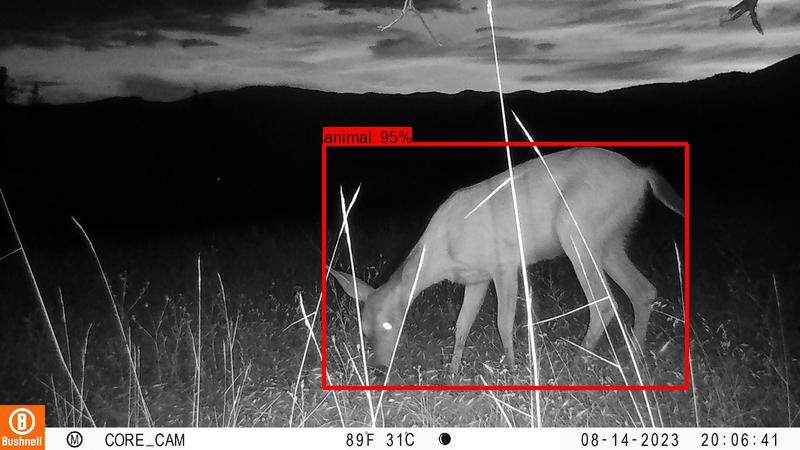}
  \caption{A black-tailed deer (\textit{Odocoileus hemionus columbianus} detected with GreenCrossingAI}
  \Description{A trail camera photo of a black-tailed deer, with red bounding box superimposed.}
  \label{fig:teaser}
\end{figure*}

\section{The Role of Camera Traps and MegaDetector for Conservation}
\begin{figure*}
    \centering
    \begin{subfigure}[b]{0.47\textwidth}
        \centering
        \includegraphics[width=\textwidth]{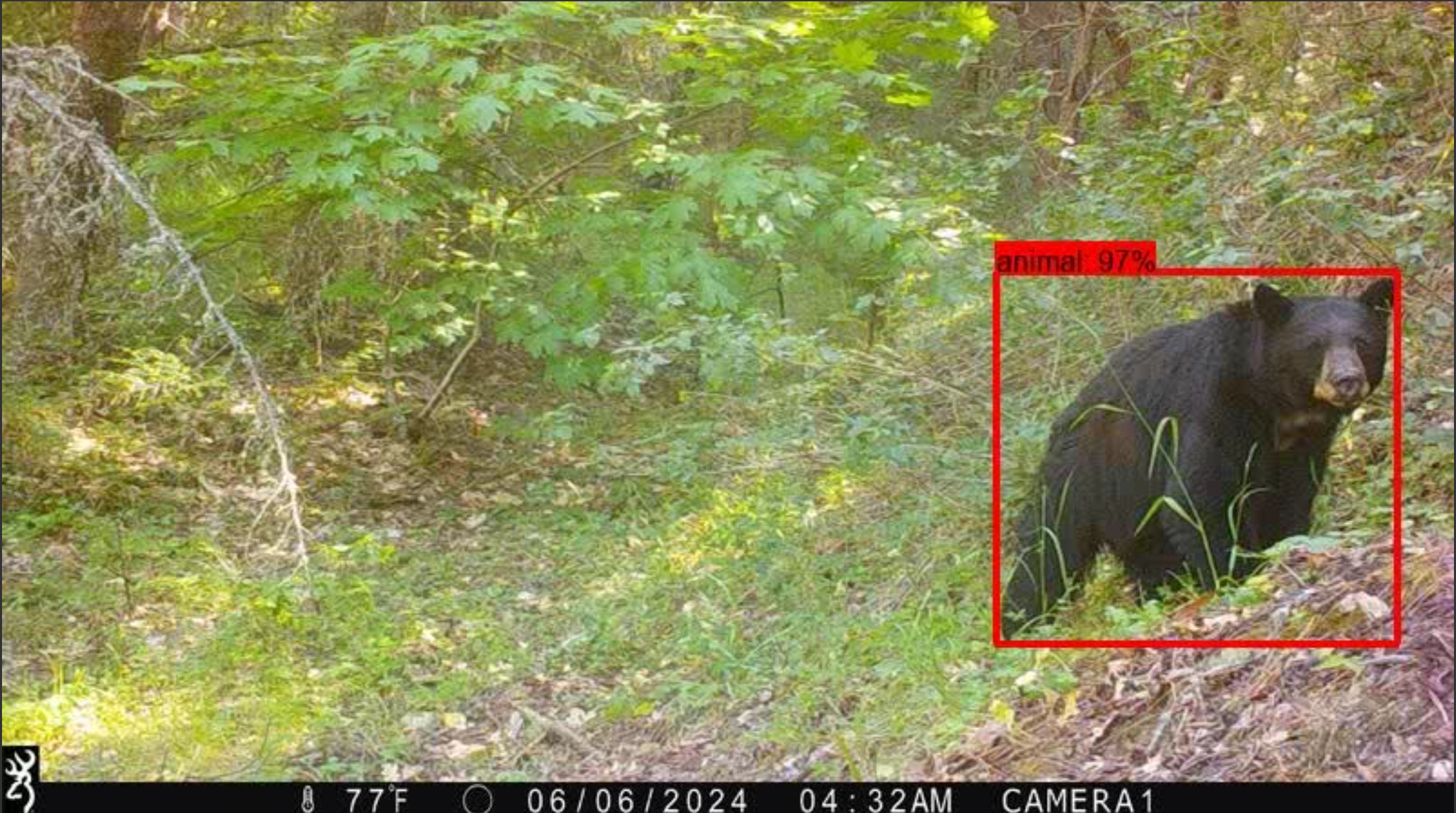}
        \caption{Black bear (\textit{Ursus americanus})}
    \end{subfigure}
    \begin{subfigure}[b]{0.47\textwidth}
        \centering
        \includegraphics[width=\textwidth]{detections_animal_08140009.JPG}
        \caption{Black-tailed deer (\textit{Odocoileus hemionus columbianus)}}
    \end{subfigure}
    \begin{subfigure}[b]{0.47\textwidth}
        \centering
        \includegraphics[width=\textwidth]{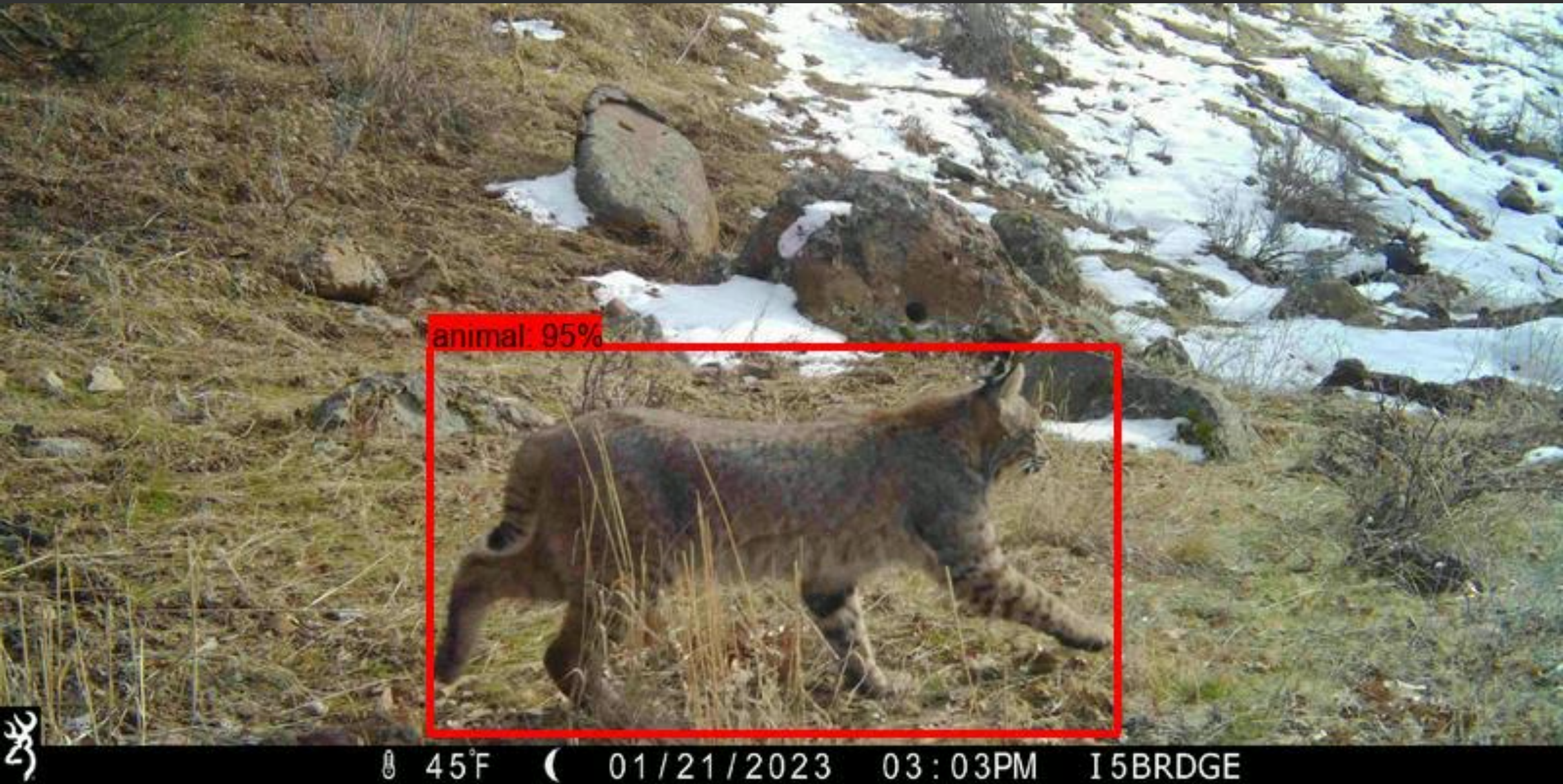}
        \caption{Bobcat (\textit{Lynx rufus})}
    \end{subfigure}
    \begin{subfigure}[b]{0.47\textwidth}
        \centering
        \includegraphics[width=\textwidth]{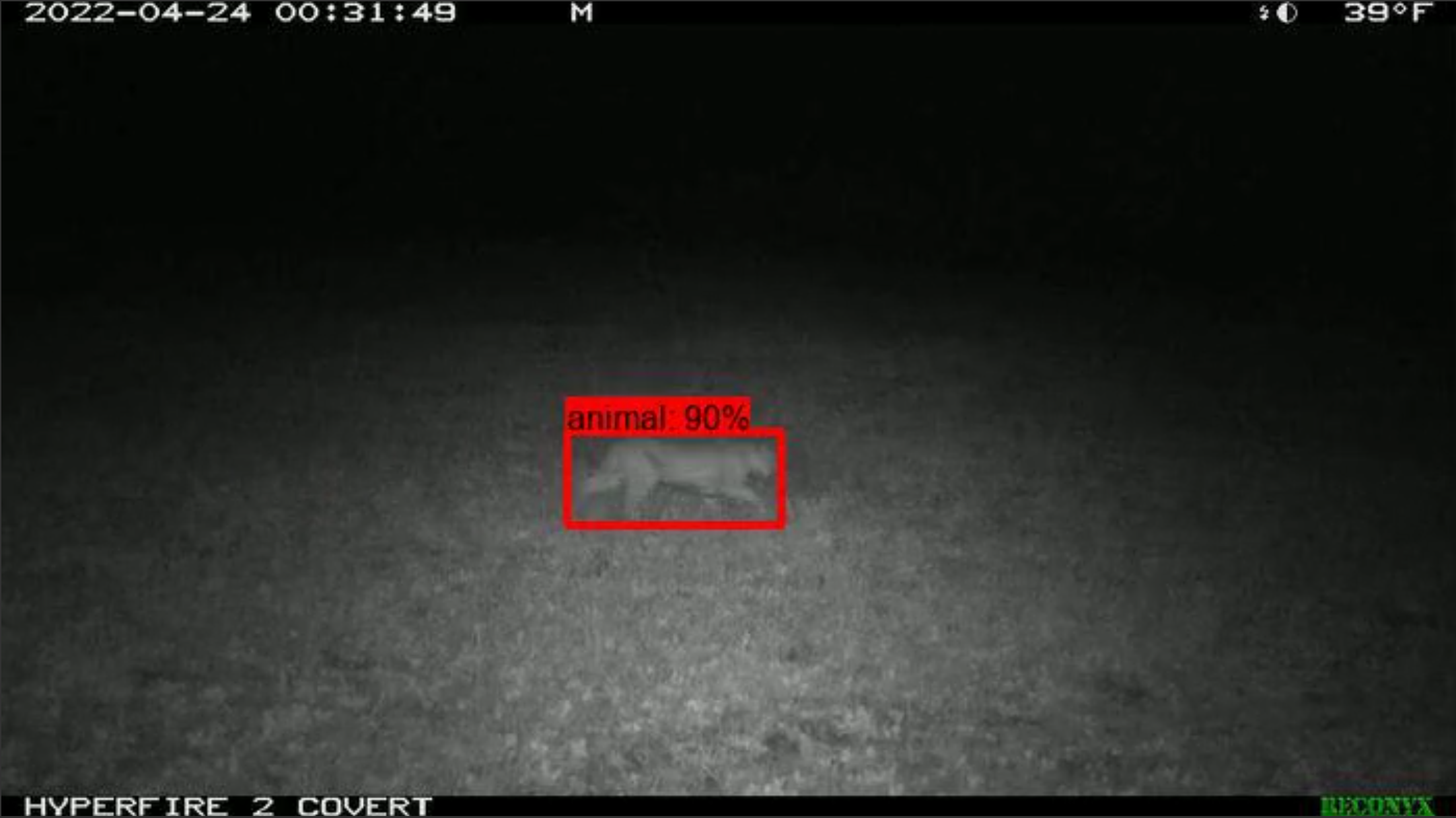}
        \caption{Bobcat (\textit{Lynx rufus})}
    \end{subfigure}
    \begin{subfigure}[b]{0.47\textwidth}
        \centering
        \includegraphics[width=\textwidth]{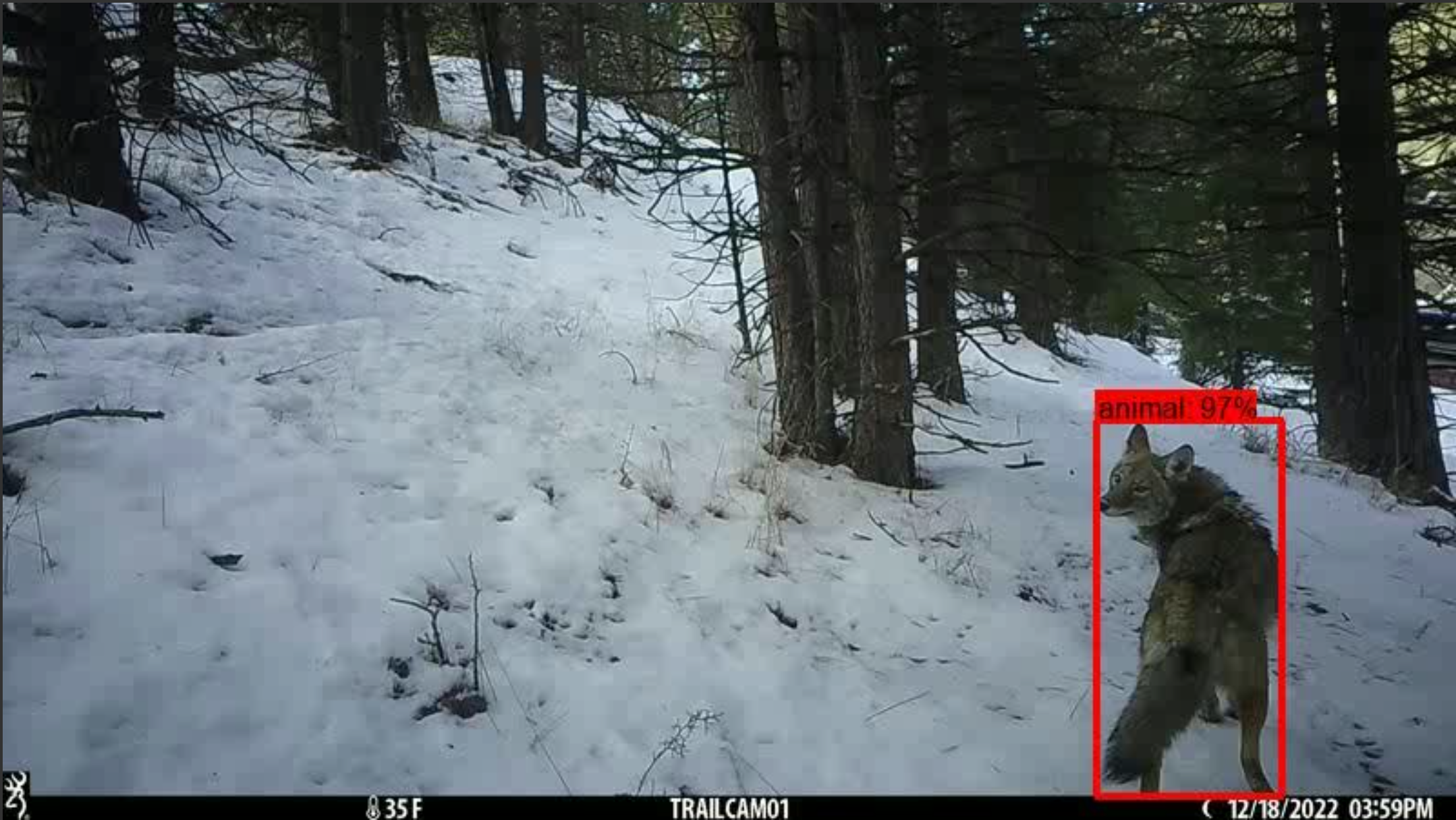}
        \caption{Coyote (\textit{Canis latrans})}
    \end{subfigure}
    \begin{subfigure}[b]{0.47\textwidth}
        \centering
        \includegraphics[width=\textwidth]{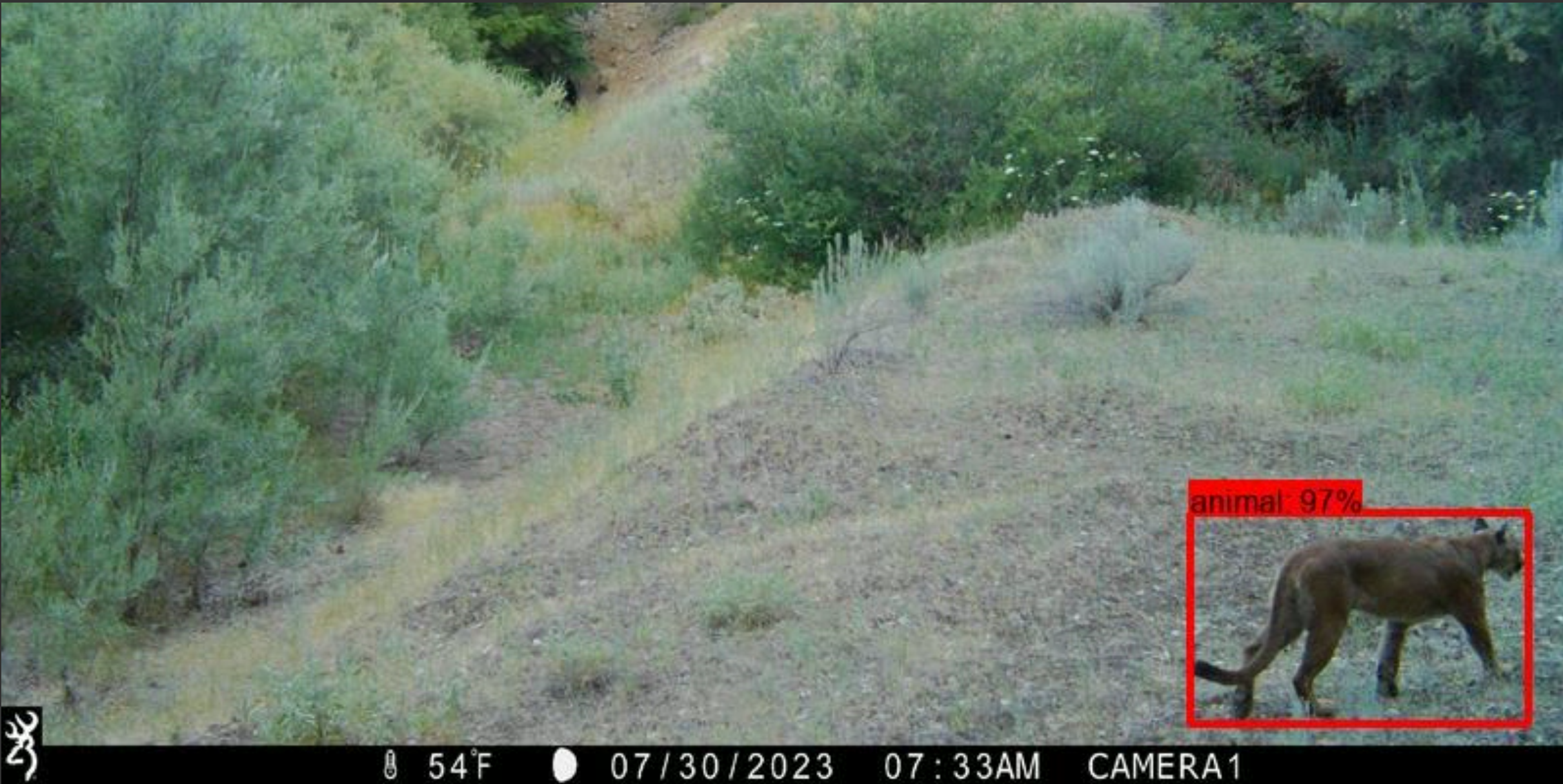}
        \caption{Cougar (\textit{Puma concolor})}
    \end{subfigure}
    \label{fig:detections}
    \caption{Wildlife detected using GreenCrossingAI.}
    \Description{Six images depicting a black bear, black-tailed deer, two bobcats, a coyote and a mountain lion detected using GreenCrossingAI, with red bounding boxes superimposed.}
\end{figure*}
Camera traps detect movement and capture images or videos in response. They function in various conditions, including infrared imaging at night or standard imaging during daylight hours, and often record additional metadata such as temperature, time, and other environmental parameters. Over recent decades, camera traps have evolved into a robust tool for monitoring wildlife and habitats, incorporating advanced features like wireless data transmission and edge computing capabilities \cite{steenweg_scaling-up_2017}. These advancements have broadened their applications in conservation, enabling more efficient and automated data collection.
MegaDetector, a machine learning model designed specifically for camera trap data, plays a pivotal role in this ecosystem. As stated by its creator Dan Morris, ``MegaDetector is an object detection model that is used to identify camera trap images that contain animals, people, vehicles, or none of the above; in practice, it’s primarily used to eliminate blank images from large camera trap surveys'' \cite{noauthor_everything_nodate}. This capability is essential for conservation projects, where the majority of captured images often contain no significant activity, and manual review would otherwise be prohibitively time-consuming.
MegaDetector versions are accessible in several open repositories and have been integrated into some online platforms to streamline its use. The model began as a YOLOv5 (You Only Look Once) based computer vision tool, optimized for speed and accuracy, that identifies and labels objects in three broad categories: animal, person, and vehicle \cite{redmon_you_2016}, \cite{noauthor_microsoftcameratraps_2024}. YOLOv5, a state-of-the-art object detection architecture, enables MegaDetector to draw bounding boxes around detected objects and label them accordingly, facilitating quick and efficient image analysis. Many researchers have added classification capabilities to MegaDetector, allowing for species labeling, with varying degrees of success.
The development of MegaDetector at Microsoft leveraged a diverse array of training data, including images from the LILA BC (Labeled Information Library of Alexandria) datasets, ensuring its adaptability across various ecological settings \cite{noauthor_data_2024}. While it is the industry standard for detecting animals in camera trap images, MegaDetector has realistic limitations. Errant objects, such as sticks or boulders resembling tortoise shells, may occasionally be misclassified. However, this tendency errs on the side of over-detection, which is preferable in wildlife research where missing critical observations could have greater consequences. This balance makes MegaDetector a highly valuable tool in conservation efforts, where accuracy and efficiency are paramount.

\section{One End Goal: The Land Bridge Project}

Numerous partner organizations, including Southern Oregon University (SOU) in Ashland, Oregon, came together to form the Southern Oregon Wildlife Crossing Coalition (SOWCC) in 2021 to address the growing need for wildlife-safe crossings over the busy I-5 freeway in southern Oregon \cite{noauthor_sowcc_nodate}. Their first major objective was a proposal to construct a wildlife overpass, providing a greenway that would significantly reduce accidents caused by animals attempting to cross the freeway \cite{heller_sou_2024}. The I-5 freeway traverses the state of Oregon and has effectively cleaved a highly-biodiverse, montane wildlife habitat in two. Animals may risk crossing the freeway–and they do–resulting in deaths to animals and humans, as well as millions of dollars of damage annually \cite{noauthor_oregon_nodate}.
The Oregon Department of Transportation has recently received  a \$33 million award from the Federal Highways Administration to build the Mariposa wildlife overpass \cite{wozniacka2024oregon}. Environmental Science (ES) students at SOU, led by Dr. Karen Mager, have undertaken data collection and analysis efforts at the proposed bridge site since 2022. The ES students previously processed the images by hand, which was naturally time consuming. The ES students went to the site, swapped out SD cards and batteries in the camera traps, brought the SD cards back to campus, searched each image for wildlife, then used Timelapse \cite{noauthor_home_2024} to label and put the images into a database stored in Box \cite{noauthor_home_2024}. 96 percent of the images do not contain an animal in them (\textasciitilde40,000 images out of over one million images contain animals), even though the camera trap was triggered by movement to take a photo. 
Computer scientist Dr. Bernie Boscoe, also at SOU, and Dr. Mager formed an ES-CS collaboration to integrate ML/AI into the pipeline, aiming to enhance efficiency while monitoring its power usage. Their ultimate goal is to transition the system onto SOU’s solar grid \cite{noauthor_green_nodate}, \cite{noauthor_greencrossingai_nodate}. 

\section{Existing Solutions}
Numerous tools and workflows exist for analyzing camera trap images, each with distinct advantages and limitations. As Velez \cite{velez_choosing_2022} highlights in a recent paper, these workflows range from CPU-only solutions, which can be run on standard machines but are extremely slow for large datasets, to cloud-based solutions, which leverage high-performance computing (HPC) to process images efficiently and return results to the user. However, the literature lacks examples of a middle-ground solution: a GPU-enabled local machine capable of processing images efficiently without the costs and complexities of cloud infrastructure yet using existing MegaDetector models and code.
At Microsoft’s PyTorch Wildlife GitHub repository, a wealth of models, implementation code, and sample use cases are freely available \cite{hernandez_pytorch-wildlife_2024}. While this resource provides powerful tools for camera trap analysis, its vast and constantly evolving repository of options can be overwhelming for newcomers, particularly those without extensive computational expertise \cite{noauthor_microsoftcameratraps_2024}. Navigating these resources requires significant time and understanding, which can be a barrier for researchers who need practical solutions that align with their research goals.
Before the GreenCrossingAI collaboration, Dr. Mager attempted to use Microsoft's cloud processing service to analyze her camera trap images. Microsoft provided the service at no cost, processing the images and returning results. However, due to the constant influx of new images, Dr. Mager felt uncomfortable relying on an external service to handle tens of thousands of images over an extended period. This dependency raised concerns about sustainability, and the logistical burden of continually uploading large datasets. These limitations prompted her to explore alternative solutions that could be implemented locally, ultimately leading to the development of the GreenCrossingAI pipeline.

\section{Requirements Gathering in Interdisciplinary Collaborations}
The development of the GreenCrossingAI pipeline was influenced by several constraints, informed by interviews with Dr. Karen Mager, Boscoe, and both Computer Science (CS) and Environmental Science (ES) capstone students. One of the primary considerations was the technical knowledge of the students. Both Mager and Boscoe sought to create a system that engaged students in the scientific process rather than relying on a fully automated, cloud-based workflow that required minimal interaction. They believed that a "push-button" solution would strip away the tactility of the scientific process and reduce the students’ agency in decision-making at various points in the pipeline. Instead, they envisioned a guided, hands-on approach that would allow students to interact meaningfully with the tools, helping them develop technical skills while maintaining ownership of their research.
Most ES students were familiar with Windows and tools such as Excel, R, and, occasionally, Jupyter notebooks. However, many expressed reluctance about learning Linux or working with command-line interfaces. Furthermore, some students voiced general concerns about artificial intelligence “taking their jobs,” preferring to spend time in the field rather than in front of a computer. Despite these reservations, the students agreed to learn how to use MegaDetector, provided the system was easy to use and produced bounding boxes that allowed them to label and classify animals efficiently.

Student availability presented another challenge, given the under\-grad\-u\-ate-only nature of the research group. Students rotate in and out of the group frequently, making onboarding and documentation essential. Mager and Boscoe serve as the project’s only consistent anchors, emphasizing the importance of clear, accessible training materials for newcomers. Additionally, ES students focus on their individual research objectives, viewing the pipeline as a tool to facilitate their projects rather than the primary focus of their efforts.

Hardware and software constraints further shaped the project. The team relied heavily on Timelapse, a Windows-only tool that plays a critical role in their labeling and tagging workflow. The students were reluctant to switch to Linux, which is typically the preferred operating system for machine learning tasks. Moreover, SOU’s Box storage system works seamlessly with Windows but is incompatible with Linux, reinforcing the decision to remain within the Windows ecosystem. Budget constraints were another significant factor; with only a few thousand dollars available, cloud-based solutions were deemed too costly. A GPU-enabled local machine was chosen as the most cost-effective option to meet the group’s needs.

The fieldwork environment introduced additional challenges. The camera traps used at the site are a mix of donated equipment of varying ages, with SD cards and batteries that require frequent replacement. The field site itself is remote, with spotty internet connectivity and extreme seasonal weather conditions, from scorching summers to cold, wet winters. These factors necessitated a system that could function reliably in challenging conditions, have data be moved to campus and stored locally for research use before archiving it on SOU’s Box system.

Privacy and security were also key considerations. Camera trap data can reveal the locations of endangered species, raising concerns about poaching and other risks. While Microsoft’s cloud service had provided a secure option for processing images in the past, there was uncertainty about whether other cloud providers would offer similar guarantees. This, combined with the logistical burden of uploading large datasets, made a local solution more appealing.
Finally, environmental concerns added another layer to the decision-making process. By developing a local solution, the team could explore integrating the system into SOU’s solar grid, aligning with the university’s sustainability goals. The rising energy demands of AI-driven cloud data centers and the lack of robust policies to curb their environmental impact further reinforced the appeal of a locally hosted pipeline.

\section{System Design}
In 2024, an Alienware Aurora R15 was purchased for \$2,750, customized with an NVIDIA GeForce RTX 4090 (24GB VRAM), an AMD Ryzen 7900X processor, and 2TB of storage. An Elmor Labs Power Measurement Device with USB (PMD-USB) was installed on the GPU, as the compact case design limited access to the CPU \cite{noauthor_elmorlabs_nodate}. A USB-C-connected SD card reader was added, along with a QNAP backup device featuring four hard drive bays, various RAID configurations, and 12TB of storage. The system was integrated into the campus network with CrowdStrike security. A custom build was not selected due to a combination of time limitations, institutional procurement policies, and the preference for a vendor-supported, pre-assembled system with warranty coverage.

The computer runs Windows 11 and has Jupyter, Timelapse, Git for Windows, and MegaDetector installed. Windows Remote Desktop enables remote access, and a shortcut to Box is available on the desktop. A monthly backup plan covers operating system configurations, while Jupyter notebooks are stored in GitHub. These notebooks include installation instructions for MegaDetector and guides for researchers on locating image folders, selecting between image and video data, naming results, and determining storage locations.

\section{Pipeline Workflow}
\subsection{Data Transfer from the Cameras}
Students visit field sites along I-5 to swap out SD cards and replace camera batteries. A full round includes approximately 28 cameras. After returning to campus, they use an SD card reader on the Alienware computer to transfer the data. They create a folder named after the collection date, such as 3May2024 with subfolders for each camera site visited on that date.

\subsection{Running MegaDetector}
\begin{table}[h!]

    \caption{Transferring images to the computer, in a typical session}
    \begin{tabularx}{\columnwidth}{Xrrr}
        \toprule
        & Min & Max & Avg\\
        \midrule
        Number of cards uploaded per session & 1 & 24 & 8\\
        Card upload time in minutes (from reader to machine) & $<$1 & 55 & 7\\
        Total GB per card & 0.5 & 119 &25\\
        \bottomrule
    \end{tabularx}
\end{table}
Next, students open the Jupyter notebook to run the batch processing. They update the cell containing the folder path for MegaDetector (MD) and specify the output locations for the resulting JSON file and the processed images with bounding boxes. MD offers various output formats, including an HTML page with categorized folders (Animal, Person, Vehicle, Other), but the SOU group does not use this feature, as they have integrated image review into their existing workflow.

Students then run MD on each camera location folder, with processing taking approximately 20 minutes per folder. In a full workday (about six hours), they can complete end-to-end labeling and processing for approximately 10 SD cards. A typical session runs as shown in Table \ref{tab:transfer-stats}.

\begin{figure*}

    \includegraphics[width=\textwidth]{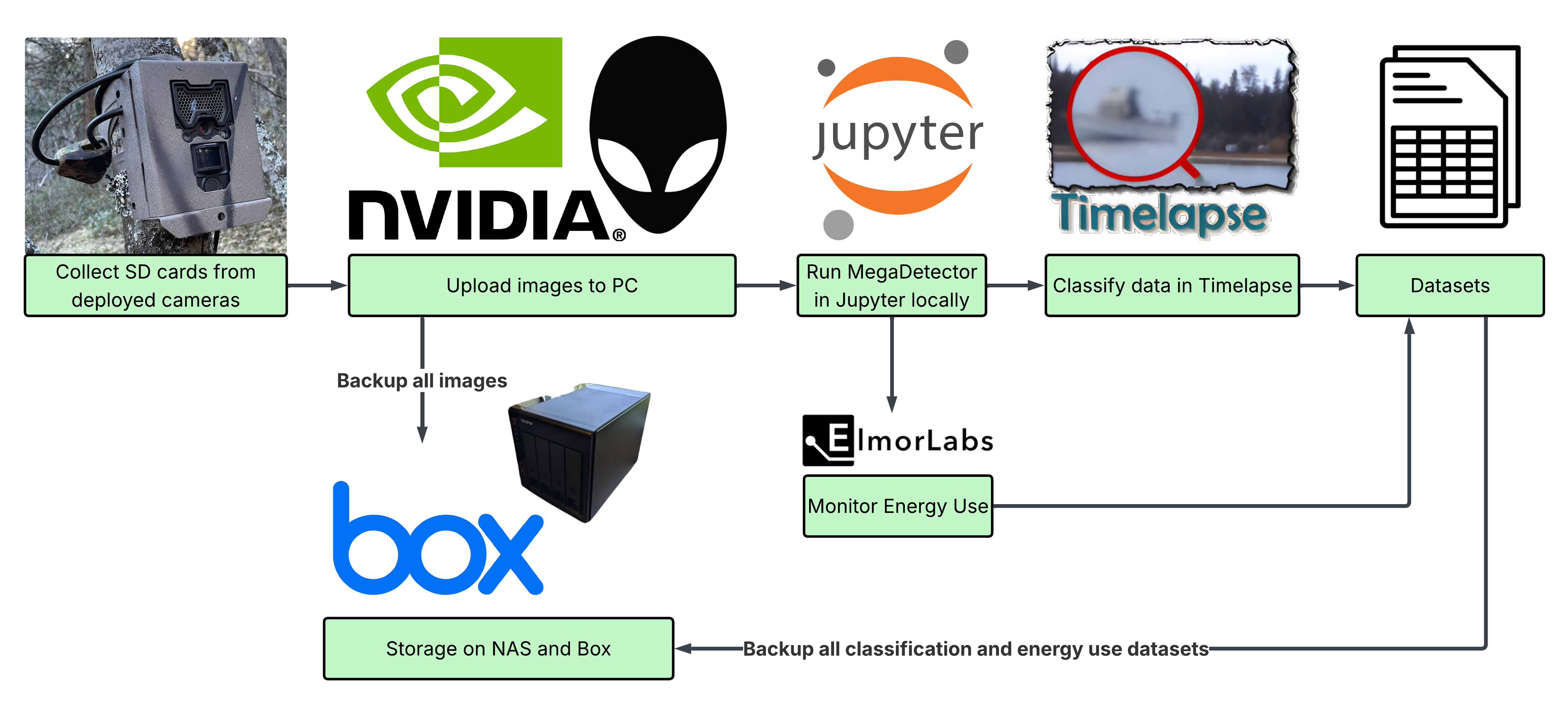}
    \caption{Flowchart depicting workflow from fieldwork to analysis-ready datasets.}
    \label{fig:flowchart}

\end{figure*}

\subsection{Running Timelapse}
\begin{figure*}[h]
\begin{subfigure}[b]{0.47\textwidth}
        \centering
        \includegraphics[width=\textwidth]{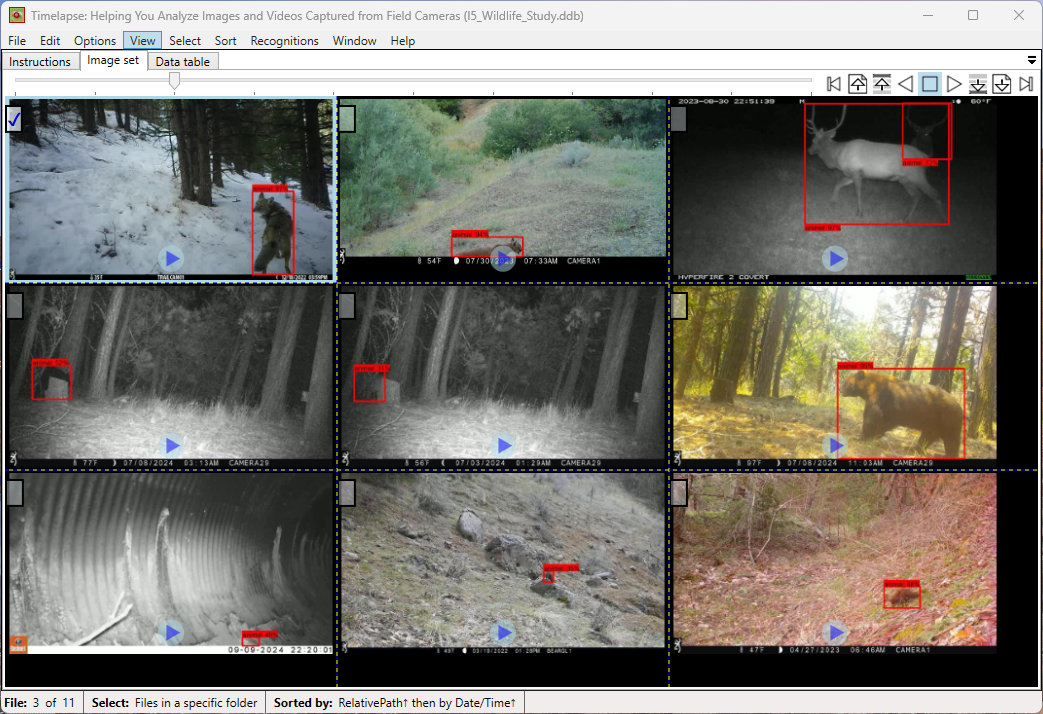}
    \end{subfigure}
    \begin{subfigure}[b]{0.47\textwidth}
        \centering
        \includegraphics[width=\textwidth]{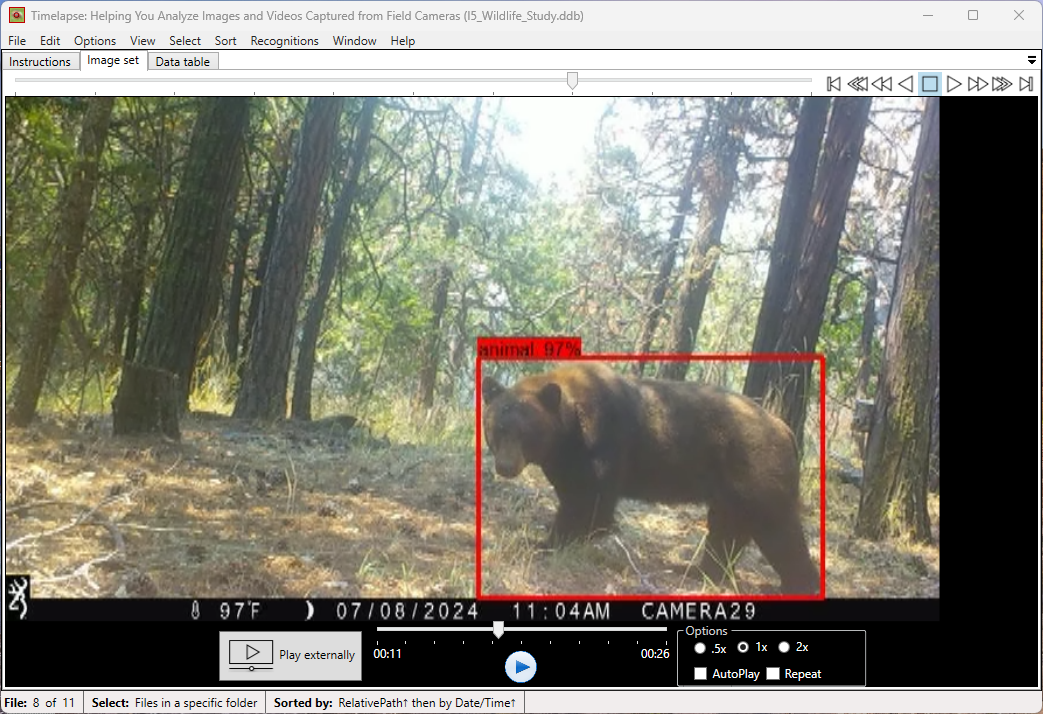}
    \end{subfigure}
    \caption{Timelapse user interface displaying detections.}
    \label{fig:timelapse}
\end{figure*}
Before MegaDetector was implemented into the pipeline, students manually reviewed images in Timelapse, clicking through each one to identify animals. When an animal was detected, they logged the image accordingly. During data cleaning, `events' were classified—if multiple images of the same animal or group appeared within a ten-minute span, they were categorized as a single event. For example, seven images of a deer group photographed within ten minutes would count as one event. With MegaDetector integrated into the workflow, students can now view a grid of image thumbnails, quickly identifying those with bounding boxes. They review these images in the context of time and location, labeling them in sequence. The same ten-minute event classification rule applies: one representative image is selected, and students log species, count, temperature, and other details in the Timelapse GUI.

Once MD results are verified, students export the Timelapse data table as a .csv file, which becomes the working dataset. Datetime values are cleaned, humans, unknowns, and birds are removed unless relevant, such as birds that might not fly far (e.g., wild turkey (\textit{Meleagris gallopavo})) for overpass analysis. From this dataset, students can conduct their own investigations, often analyzing specific species or locations using tables and statistical tools like R.
\subsection{Storing Results}
The images processed by MegaDetector are stored in designated folders on Box, along with Timelapse databases, CSV files, and student analyses. Due to slow upload speeds, image transfers to Box are typically run as overnight jobs. When timed during the day, the total data uploaded to Box was 2.2 terabytes. Table \ref{tab:file-stats} shows summary statistics about upload times to Box.
\begin{table}[h]
\caption{File Size and Upload Time Statistics}

\centering
\begin{tabular}{lr}
\hline
Metric & Value \\
\hline
Minimum file size & 0.05 GB \\
Maximum file size & 221.00 GB \\
Average file size & 28.70 GB \\
\hline
Total upload time & 344 minutes \\
Minimum upload time & $<$1 minute \\
Maximum upload time & 50 minutes \\
Average folder upload time & 4.6 minutes \\
\hline
\end{tabular}
\label{tab:file-stats}
\end{table}

Previously, transferring data from an SD card on a 2015 MacBook Pro could take 4–5 hours for a single 32GB card, making the Alienware system a significant improvement in data upload speed.
\section{Results of Implementation}
Since 2021, 20–35 camera traps have collected 6.7TB of data, consisting of approximately 1 million photos and videos. Of these, \textasciitilde
40,000 contain animals, representing 8,295 unique events and 11,624 individual animals. If an animal remains in the camera’s field of view for multiple detections, it is counted as a single individual.

Following the implementation of MegaDetector (MD), 10,000 images were processed through MD and compared against original human labels, with a computer science student independently verifying the results. MD is designed to err on the side of caution, occasionally detecting swaying branches, rocks, or trees that resemble antlers, resulting in a small number of false positives. It can also miss detections within a series—for example, identifying only one frame in a sequence where a deer walks by or incorrectly boxing two deer behind a tree as three individuals. However, students using Timelapse can easily catch and correct these errors. Since at least one frame in a sequence typically contains a correct detection, the percentage of missed animals remains very low.

MD is still in the exploratory phase of implementation, and further refinements are ongoing. Since the bridge project has been approved, these results will serve as an excellent baseline for continued observations of animals in the area. The long-term dataset allows researchers to monitor shifts in wildlife activity before, during, and after construction, providing valuable insights into how species respond to habitat changes. With MegaDetector now integrated into the workflow, processing new data will be significantly more efficient, enabling faster detection and analysis of movement patterns, species distribution, and potential disturbances.

Future refinements to the detection pipeline, including improved filtering of false positives and enhanced classification models, will further streamline the workflow. Additionally, the ability to compare MD results against human annotations will continue to refine best practices for automated wildlife monitoring. As more data is collected, researchers will also have the opportunity to analyze seasonal trends, species interactions, and long-term ecological changes, contributing to both conservation efforts and informed decision-making regarding land use and infrastructure development in the region.
\section{Maintenance}
Maintaining the system and pipeline over time will require a combination of hardware upkeep, software updates, and workflow refinements to ensure continued efficiency and accuracy. Regular system maintenance, such as updating MegaDetector, Jupyter notebooks, and Timelapse, will be necessary to keep compatibility with evolving machine learning models and operating systems. Hardware components, including the Alienware workstation, SD card readers, and storage devices, will need periodic checks to prevent failures and ensure optimal performance. Data management will remain a key priority, with backups of processed images, Timelapse databases, and analysis files stored in Box and external drives to safeguard against data loss. As the datasets grow in size, findability and discoverability at scale become increasingly important to ensure that researchers can efficiently access, analyze, and share data. Establishing a well-structured metadata system, including standardized naming conventions, timestamps, location tags, and species annotations, will help streamline search and retrieval. Leveraging indexing tools, database queries, or even AI-driven search capabilities could further enhance usability, allowing researchers to filter data based on specific parameters such as date ranges, species presence, or environmental conditions.

Additionally, integrating cloud-based or institutional data repositories with clear documentation will improve long-term accessibility and collaboration, ensuring that datasets remain useful beyond the immediate project. As new methods for large-scale ecological data management emerge, periodic assessments of storage solutions and retrieval mechanisms will be necessary to maintain efficiency and support expanding research questions. Additionally, periodic reviews of MegaDetector’s performance, comparing automated detections with human annotations, will help refine detection thresholds and reduce false positives or missed identifications. As new students join the project, clear documentation and onboarding materials will be essential to ensure continuity in data collection and processing. Over time, iterative improvements to the workflow, including automation of repetitive tasks and potential integration of cloud-based processing, may further enhance scalability and long-term sustainability of the system.
\section{Future Work}
In general, questions abound about how to improve MegaDetector for the environmental scientists' needs. Fine-tuning MegaDetector could entail transfer learning, adjusting thresholds to prioritize recall over precision and reduce missed detections. Another option is to build a classifier on top of MegaDetector. As it stands, MegaDetector works well enough to save time by boxing images, and since the team consists of environmental science experts, they do not require precise automated classifications. Instead, they prefer to manually choose and label images as part of their workflow. However, the persistent issue of empty images filling up the camera trap storage out in the field and taking up hard drive space on drives remains a challenge. As a result, the group closely follows new studies on MegaDetector and considers how they might improve their pipeline. Recent advances include a study by Gadot et al. \cite{Gadot}, which explores whether species classification is more effective using cropped images (bounding box extractions) or whole images. Their findings indicate that cropped images improve species classification accuracy by reducing background noise and emphasizing the detected animal. While whole images retain environmental context that might sometimes aid classification, the study suggests that focusing on cropped images leads to better performance overall. These insights are valuable for refining the use of MegaDetector, as they indicate that integrating object detection into the classification pipeline can enhance accuracy and streamline species identification. This research will help optimize how the group processes and analyzes growing datasets in their workflow.
These insights are valuable for refining the use of MegaDetector, as they indicate that integrating object detection into the classification pipeline can enhance accuracy and streamline species identification. This research will help optimize how the group processes and analyzes growing datasets in their workflow.

Other possibilities for new directions include addressing location-specific overfitting, for example, looking at why certain rocks or branches end up being bounded-boxed as animals. In this case, location-specific overfitting refers to a situation where a machine learning model, like a classifier used for camera trap images, becomes too tailored to the specific characteristics of images from a particular location. The model might learn to recognize patterns that are unique to that location—such as certain backgrounds, lighting conditions, or environmental factors—rather than focusing on the actual animals or species being detected.

As a result, when the model is applied to images from different locations, its performance might decrease because it has overfitted to location-specific features that are not present in the new location. This leads to poor generalization of the model across different environments. 
In their study "Towards Zero-Shot Camera Trap Image Categorization," Vyskočil et al. \cite{vyskocil_towards_2024} explore methods to enhance automatic categorization of camera trap images and mitigate location-specific overfitting. They evaluate approaches that combine MegaDetector with one or more classifiers, as well as the Segment Anything Model (SAM). Their findings indicate that integrating MegaDetector with two separate classifiers achieves the highest accuracy, reducing relative error by approximately 42\% on the CCT20 dataset, 48\% on the CEF dataset, and 75\% on the WCT dataset. Additionally, by removing background information, the error in terms of accuracy in new locations is reduced by half. These results suggest that combining MegaDetector with multiple classifiers and employing segmentation techniques can effectively reduce location-specific overfitting, leading to more accurate and generalizable species identification in camera trap images.

Looking ahead, there is also interest in exploring how machine learning could support other aspects of environmental science research beyond species classification. For example, ML models might assist with behavior recognition, anomaly detection, or identifying changes in habitat over time. Even when manual labeling remains preferred for core classification tasks, ML could still play a role in prioritizing data for review, filtering out false positives, or flagging unusual patterns that merit further investigation. These possibilities suggest that, as datasets scale, the integration of machine learning into the research workflow may become increasingly valuable—not to replace expert judgment, but to amplify it.

\section{Conclusion}
In this paper, we demonstrated the implementation of a local pipeline using MegaDetector on an Alienware machine to streamline the processing of camera trap data for wildlife monitoring. By integrating MegaDetector into our workflow, the group significantly reduced the time and effort required for image labeling, allowing the team of environmental science experts to focus on the analysis and interpretation of the data. We discussed the many challenges in meeting the needs of environmental scientists as they incorporate modern technological tools into their workflows. Future work for the entire collaboration will focus on developing a low-power energy solution, investigating ways to handle empty images, and addressing issues such as managing large datasets and reducing storage constraints. Future research done by computer science students might entail projects such as improving detection accuracy, or introducing a classification layer. Overall, this pipeline represents a significant step forward in our ability to efficiently handle and analyze large volumes of wildlife camera trap data for a small environmental science research group, and will play an important role in advancing ecological and conservation research.

\begin{acks}
Research made possible with support from the Alfred P. Sloan Foundation, grant number: G-2024-22720 \\
Southern Oregon University's Institute for Applied Sustainability \\
Southern Oregon Wildlife Crossing Coalition \\
The Oregon Wildlife Foundation \\
WildTech Award from Wildlabs and ARM
\end{acks}

\bibliographystyle{ACM-Reference-Format}
\bibliography{pearc25}

\appendix

\end{document}